\documentclass{svjour3}
\usepackage{graphicx}

\usepackage{lscape}
\usepackage{url}
\usepackage[]{graphicx}
\usepackage[]{geometry}
\usepackage{amsmath}
\usepackage{setspace}

\usepackage{algorithm}
\usepackage{algorithmic}

\usepackage{epsfig}
\usepackage{multirow}
\usepackage{subfigure}
\usepackage{lineno}

\usepackage[numbers]{natbib}

\begin{document}

\title{OntoWind: An Improved and Extended Wind Energy Ontology}

\author{Dilek K\"u\c{c}\"uk         \and
        Do\u{g}an K\"u\c{c}\"uk
}

\institute{D. K\"u\c{c}\"uk \at
              Electrical Power Technologies Group, T\"UB\.ITAK Energy Institute, Ankara{--}Turkey\\
           \and
           D. K\"u\c{c}\"uk \at
              Department of Computer Engineering, Gazi University, Ankara{--}Turkey\\
}

\date{}

\maketitle

\begin{abstract}
Ontologies are critical sources of semantic information for many application domains. Hence, there are ontologies proposed and utilized for domains such as medicine, chemical engineering, and electrical energy. In this paper, we present an improved and extended version of a wind energy ontology previously proposed. First, the ontology is restructured to increase its understandability and coverage. Secondly, it is enriched with new concepts, crisp\slash fuzzy attributes, and instances to increase its usability in semantic applications regarding wind energy. The ultimate ontology is utilized within a Web-based semantic portal application for wind energy, in order to showcase its contribution in a genuine application. Hence, the current study is a significant to wind and thereby renewable energy informatics, with the presented publicly-available wind energy ontology and the implemented proof-of-concept system.
\end{abstract}

\section{Introduction}\label{sec:intro}

It is commonly acknowledged that domain ontologies are significant sources of semantic information \cite{Perez2004}. Hence, new ontologies are
being constructed for many different domains and they are effectively utilized within relevant applications in these domains. To name a few, in
\cite{biotop}, a domain ontology for molecular biology is proposed. In order to perform public health surveillance, an ontology for this domain
is created and described in \cite{Kucuk2017}. A domain ontology is presented in \cite{construction} for the processes in infrastructure and
construction. In \cite{materials}, an ontology for the domain of materials science and engineering is described. A domain ontology for chemical
process engineering is proposed in \cite{ontocape}. Regarding software systems, a Web service modeling ontology to describe all aspects of Web
services is presented in \cite{webont}, an ontology for scientific software metadata is proposed in \cite{gil2015ontosoft}, and an ontology to
model the variability in software product modeling is presented in \cite{kumbang}.

The domain of energy is similarly a fruitful domain for semantic applications. For different applications in the energy domain, the conceptual
modeling of the relevant energy subdomain is required and this phase will extensively benefit from domain ontologies. Similarly, ontologies will
also alleviate the interoperability problems between different applications. Yet, there are few examples of domain ontologies regarding this
domain and its subdomains. For instance, in \cite{kucuk2010pqont}, a domain ontology to model the electrical power quality parameters and events
is described. In \cite{kucuk2014wont}, a domain ontology for wind energy which was constructed through a semi-automated procedure is presented. A
high-level covering ontology for the domain of electrical energy is presented in \cite{kuccuk2015feeont} where this ontology was also aligned
with the aforementioned electrical power quality and wind energy ontologies. An ontology for energy efficiency in smart grid neighborhoods is
presented in \cite{enersip}, and finally an ontology matching system for the smart grid domain is described in \cite{smartgrid} which aims to
reduce the related interoperability issues.

In this paper, we present an improved and extended version of the wind energy ontology proposed in \cite{kucuk2014wont}, in order to increase its
understandability, coverage, and usability. We first reorganize the existing ontology hierarchy, add new concepts, attributes, and instances, in
order to arrive at a more useful and extended wind energy ontology. We make the ontology publicly-available and utilize it within a Web-based
semantic portal application for wind energy. The rest of the paper is organized as follows: In Section \ref{sec:ont}, the extended and improved
wind energy ontology is described. Section \ref{sec:system} presents the semantic portal application in which the ultimate ontology is used and
finally, Section \ref{sec:conc} concludes the paper with a summary and pointers to future work.

\section{Extended and Improved Wind Energy Ontology}\label{sec:ont}
\subsection{Initial Version of the Wind Energy Ontology}
\begin{figure}
\center \scalebox{0.55}
{\includegraphics[angle=90]{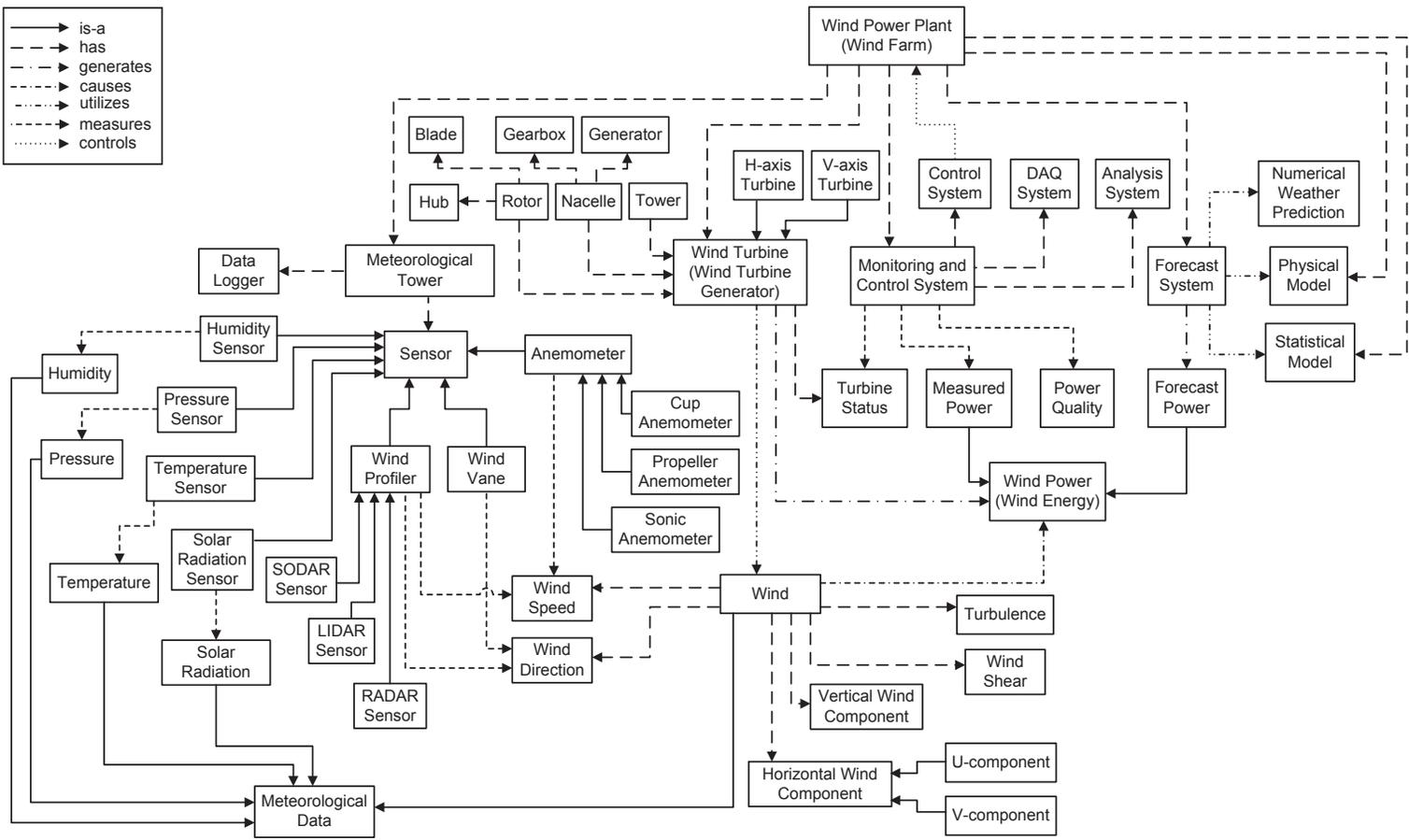}}\caption{The Initial Version of the Domain Ontology for Wind Energy \cite{kucuk2014wont}, i.e.,
WONT.}\label{fig:wont1}
\end{figure}

The initial wind energy ontology has been created through a semi-automatic procedure over Wikipedia articles related to wind energy
\cite{kucuk2014wont}. In the first phase of this procedure, related Wikipedia articles are automatically processed to determine the
high-frequency ngrams from the articles, and in the second phase, the resulting ngrams are manually organized into a wind energy ontology. The
schematic representation of the concepts of this ultimate ontology with their interrelationships is given in Figure \ref{fig:wont1} as excerpted
from \cite{kucuk2014wont}. The ontology has also been publicly shared for research purposes at
\url{http://www.ceng.metu.edu.tr/~e120329/wont.owl} as a Web Ontology Language (OWL) file.

In order to include it within the larger electrical energy ontology with weighted attributes \cite{kuccuk2015feeont}, the ontology has been
extended to include weighted attributes and this extended version is again publicly shared at
\url{http://www.ceng.metu.edu.tr/~e120329/FWONT.owl}. This form of the ontology is henceforth referred to as WONT due to the name of its initial
OWL file and the electrical energy ontology \cite{kuccuk2015feeont} is henceforth referred to as FEEONT, based on the name of its
publicly-available OWL file at \url{http://www.ceng.metu.edu.tr/~e120329/FEEONT.owl}.

\subsection{Improved and Extended Wind Energy Ontology}
In the current study, we have improved and extended WONT in order to make it more useful for applications related to wind energy. The final
improved and extended wind energy ontology is henceforth referred to as OntoWind (ONTOlogy for WIND energy).

All of the extensions and improvement efforts are carried out using the Prot\'eg\'e ontology editor \cite{protege} and the commonly-employed
ontology development methodology described in \cite{noy2001ontology} is roughly followed. In the following subsections, the main procedures
employed to construct OntoWind based on WONT are described.

\begin{figure}
\center \scalebox{0.57}
{\includegraphics{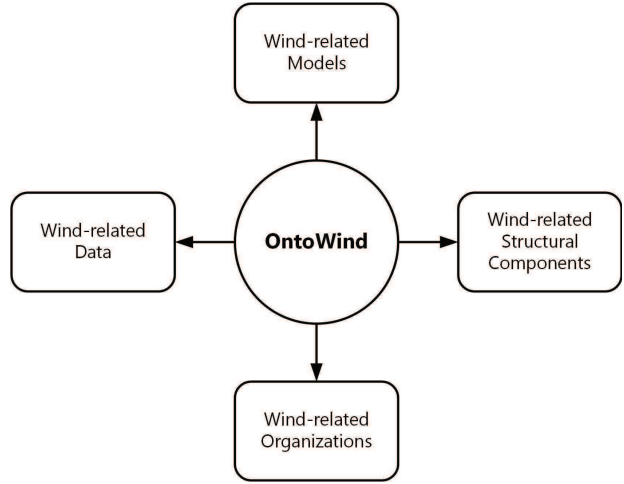}}\caption{The Most General Concepts in OntoWind.}\label{fig:wont2_genel}
\end{figure}

\subsubsection{Restructuring}
The main concepts in WONT are restructured to better separate the different concept groups. In WONT, the concept hierarchy tree was a rather
short one, as the concept-subconcept (or, class-subclass) relations were not exhaustively specified. By taking a top-down approach, we first
determine the top general concepts of the wind energy domain and the resulting four general concepts in OntoWind are illustrated in Figure
\ref{fig:wont2_genel}. Next, the existing concepts in WONT are re-modeled by putting them under their relevant general concepts. Along the way,
we exclude some concepts like the \emph{PowerQuality} which was included in WONT to model the power quality characteristics of wind energy, as
OntoWind can be integrated with external power quality domain ontologies like PQONT \cite{kucuk2010pqont} to model these characteristics.

\begin{figure}
\center \scalebox{0.63}
{\includegraphics{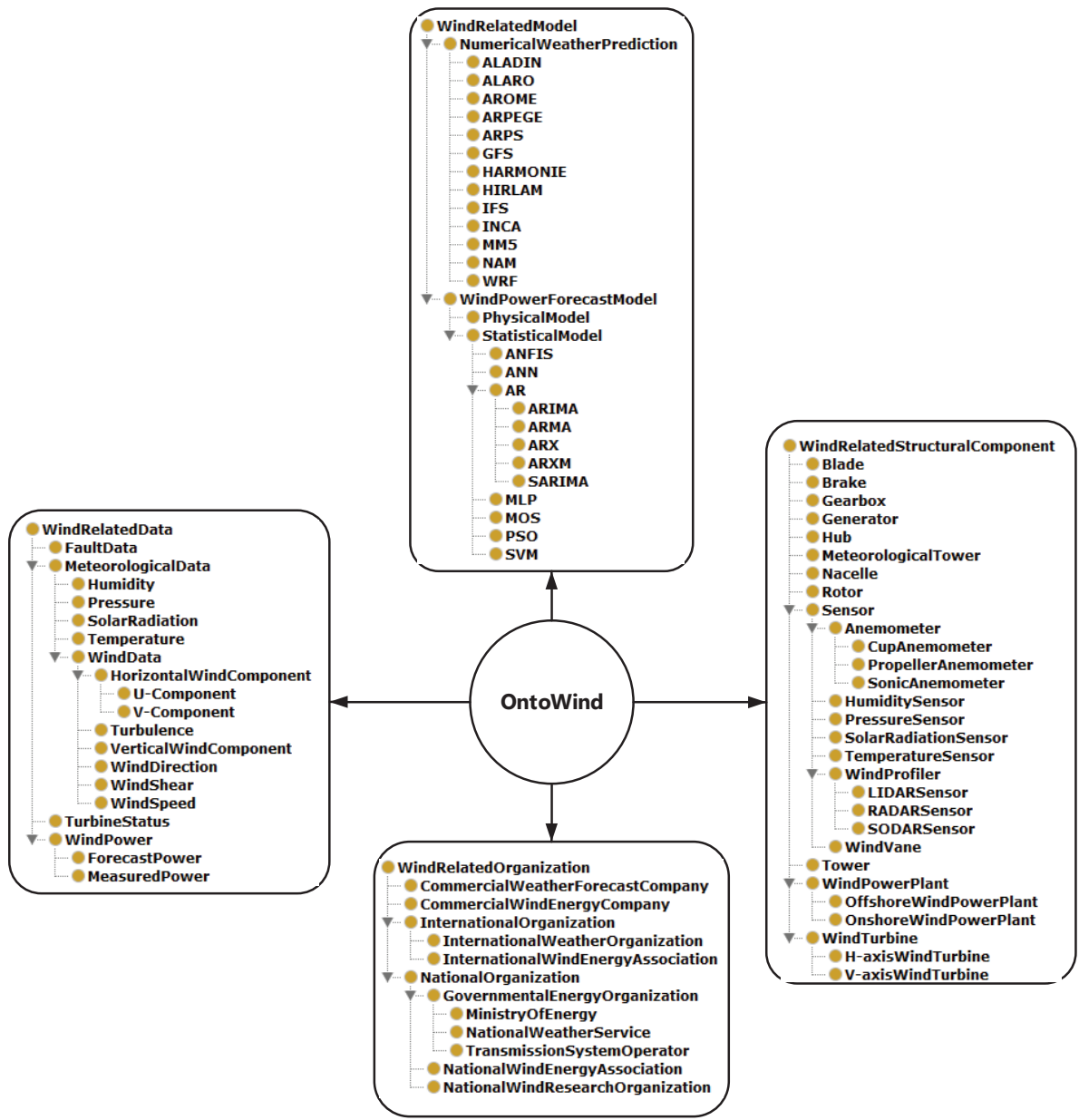}}\caption{The Concept Hierarchies under the General Concepts in OntoWind.}\label{fig:wont2_ayrinti}
\end{figure}

The ultimate concept hierarchies grouped under the four general concepts of OntoWind are provided schematically in Figure
\ref{fig:wont2_ayrinti}.

The top general concepts of OntoWind are described below:

\begin{itemize}
  \item \emph{WindRelatedData}: This concept and its subconcepts given as a tree structure in Figure \ref{fig:wont2_ayrinti} are used to model
      the measurement and forecast data related to wind energy. For instance, meteorological forecasts obtained by running the numerical
      weather prediction (NWP) models as well as meteorological measurement outputs through the related sensors are all represented by the
      corresponding meteorological data concepts included in the ontology. Namely, these concepts include \emph{WindSpeed},
      \emph{WindDirection}, \emph{Temperature}, \emph{Humidity}, in addition to others. The main attributes of these concepts are \emph{value}
      corresponding to the actual measurement\slash forecast data and \emph{date} corresponding to the actual time that the data value belongs
      to.

  \item \emph{WindRelatedModel}: This concept represents the wind-related NWP and wind power forecast models. Hence, it has the corresponding
      concepts of \emph{NumericalWeatherPrediction} and \emph{WindPowerForecastModel} for these models. These subconcepts, in turn, has the
      related commonly-used NWP models such as ALADIN, IFS, and WRF, and wind power forecast models such as ANFIS, ANN, and SVM as their
      subconcepts\footnote{The following are the open forms of the abbreviations used in the concept names:\\ALADIN: Aire Limit\'ee Adaptation
      dynamique D\'eveloppement InterNational\\IFS: Integrated Forecast System\\WRF: Weather Research and Forecasting\\ANFIS: Adaptive
      Neuro-Fuzzy Inference System\\ANN: Artificial Neural Network\\SVM: Support Vector Machine}.

  \item \emph{WindRelatedStructuralComponent}: This concept represents the physical components within a wind power plant (WPP). Hence, it has
      subconcepts such as \emph{WindPowerPlant}, \emph{WindTurbine}, and \emph{Sensor} as its subconcepts. The concepts in this category have
      the necessary object attributes to represent the details of the corresponding structural components. For instance, \emph{WindPowerPlant}
      has the related attributes to model the characteristics of WPPs such as its geo-referenced location information, installed capacity,
      number of turbines, etc.

  \item \emph{WindRelatedOrganization}: This concept represents the international and national organizations, and commercial companies related
      to wind energy. Therefore, it has subconcepts such as \emph{InternationalOrganization} and \emph{NationalOrganization} to model those
      organizations with finer granularity. Common organizations such as research centers, national weather services, and electricity
      transmission system operators are all modeled with the concepts under this category.
\end{itemize}

\subsubsection{Extensions}
Apart from the general restructuring efforts to arrive at OntoWind, it also includes several extensions over WONT.

First of all, new relevant domain concepts are added to the ontology to increase its coverage. For instance, in order to model the national and
international organizations related to wind energy, several concepts are added under the generic concept of \emph{WindRelatedOrganization}, as
previously described. Similarly, several common models for numerical weather prediction and wind power forecasting are included as subconcepts
under the generic concept of \emph{WindRelatedModel}.

Secondly, several annotation attributes are added to the ontology to make it applicable in different semantic applications. For instance, we have
added the textual attributes of \emph{webAddress} and \emph{twitterAccount} especially for the prospective instances of the subconcepts of
\emph{WindRelatedOrganization}. The former attribute will be used to model the Web site of the wind-related energy organization, while the latter
is used to hold the address of its official Twitter account, if any. Similarly, another attribute named \emph{country} is added to hold the
country codes of the national wind-related organizations.

\begin{figure}
\center \scalebox{0.71}
{\includegraphics{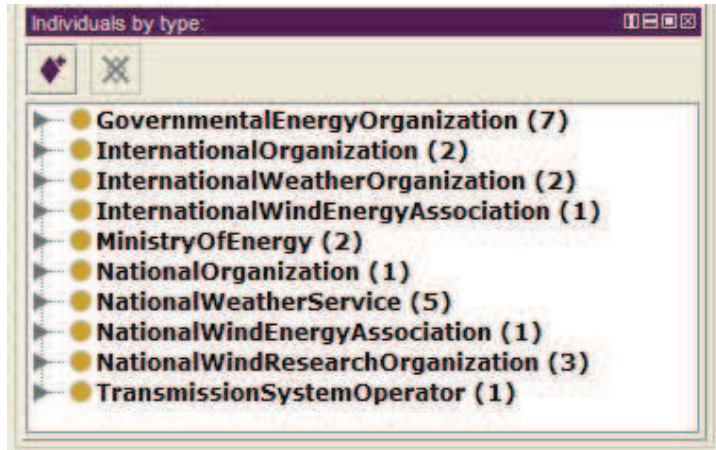}}\caption{Number of Instances per Concept Names in OntoWind.}\label{fig:instances-by-type}
\end{figure}

Finally, plausible instances (also called objects or individuals) are included in OntoWind to again make it a useful resource for Semantic Web
applications. A total of 25 instances, belonging to the subconcepts modeled under \emph{WindRelatedOrganization}, are added to OntoWind. The
numbers of instances per their immediate concept types are given in Figure \ref{fig:instances-by-type}.

\begin{figure}
\center \scalebox{0.71}
{\includegraphics{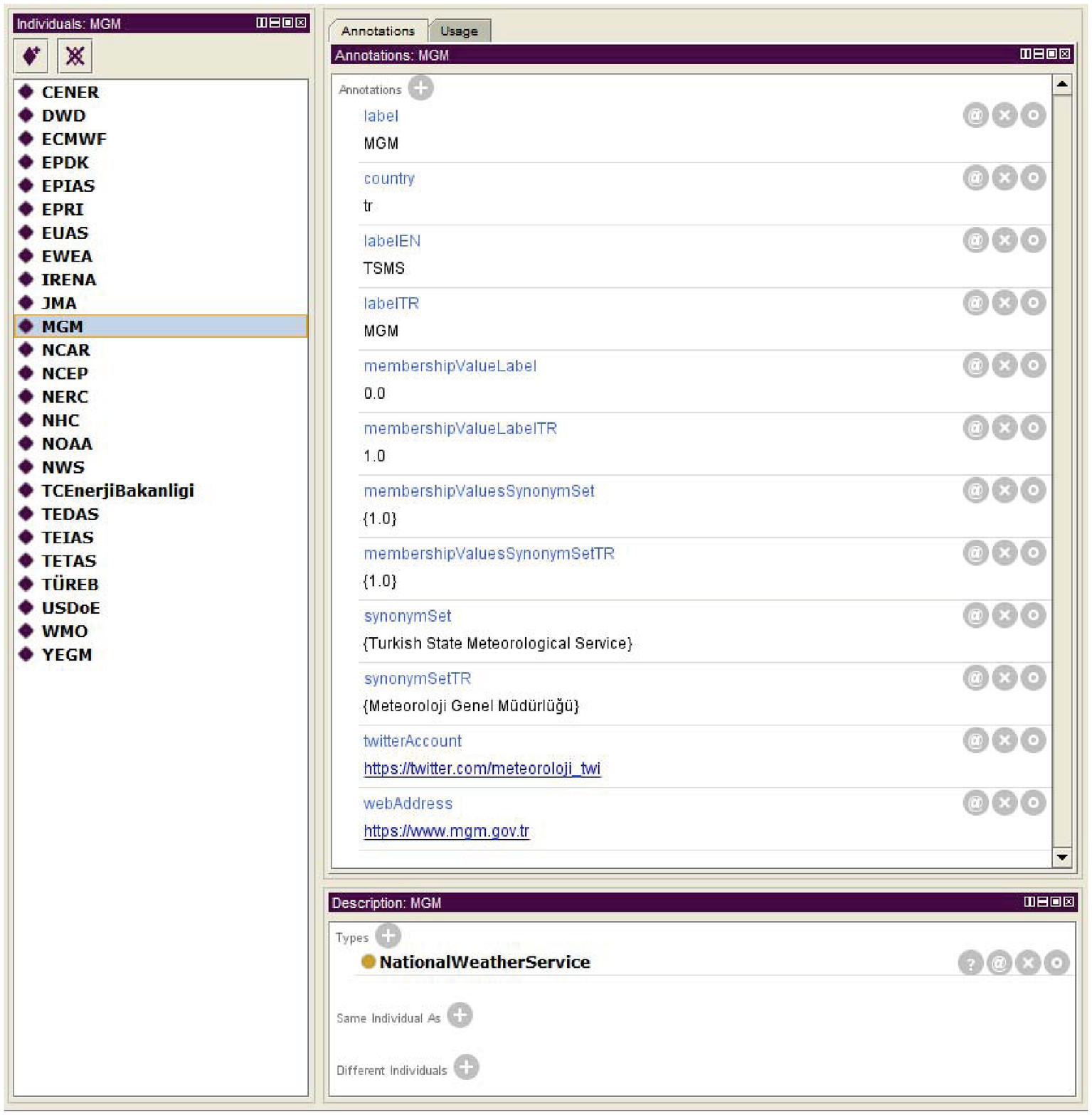}}\caption{Instances of the Subconcepts under \emph{WindRelatedOrganization} in
OntoWind.}\label{fig:instances}
\end{figure}

The names of the instances and the attributes of one of them, namely, \emph{MGM} denoting \emph{Turkish State Meteorological Service}, are
illustrated in Figure \ref{fig:instances}. MGM is an instance of the \emph{NationalWeatherService} concept which is a subconcept of
\emph{GovernmentalEnergyOrganization} and it, in turn, is a subconcept of \emph{NationalOrganization} subconcept of
\emph{WindRelatedOrganization}. The readers are referred to the concept hierarchies provided in Figure \ref{fig:wont2_ayrinti} for a schematic
view.

Figure \ref{fig:instances} also demonstrates the use of the aforementioned new attributes of OntoWind, namely, \emph{country}, \emph{webAddress},
and \emph{twitterAccount}. The remaining ones are the already existing attributes from WONT \cite{kucuk2014wont} and FEEONT
\cite{kuccuk2015feeont}, and they are explained below again, for the purposes of completeness:

\begin{itemize}
    \item   \emph{label}: This attribute holds the exact concept name.
    \item   \emph{labelEN}: This attribute holds the most common English phrase referring to the concept.
    \item   \emph{membershipValueLabel}: This attribute holds the degree of membership of the value in \emph{labelEN} to the domain of wind
        energy. The degree of membership is a real number between [0,1]. This attribute and similar attributes have previously been utilized to
        denote the weights of the corresponding attributes (like \emph{labelEN}) within FEEONT \cite{kuccuk2015feeont}. In
        \cite{kuccuk2015feeont}, Wikipedia's disambiguation pages have been used to determine the weights. In this study, we manually determine
        these weights based on expert judgement instead of the order in the Wikipedia's disambiguation pages. In \cite{kuccuk2015feeont}, if
        the sense related to wind energy is in the $n^{th}$ place among the entries in the disambiguation page corresponding to the value of
        \emph{labelEN}, then the value of \emph{membershipValueLabel} is set to $1/n$. This is the \emph{reciprocal rank} metric used in
        information retrieval and question answering literature \cite{Voorhees1999}. As previously mentioned, we have used expert judgement to
        manually determine the value of \emph{membershipValueLabel} as a real number between [0,1]. The values of all attributes beginning with
        \emph{membershipValue} are similarly determined. Also, the existing values in WONT for such attributes are revised in case a change of
        value is needed.
    \item   \emph{labelTR}: This attribute holds the most common Turkish phrase referring to the concept. As it has been pointed out in related
        work like \cite{kucuk2010pqont} and \cite{kuccuk2015feeont}, attributes like \emph{labelTR} facilitate the use of the ontology in
        multilingual settings. For instance, to extend the ontology to other languages like Spanish or French, attributes like \emph{labelES}
        and \emph{labelFR} can be added to the ontology with the corresponding values in these languages.
    \item   \emph{membershipValueLabelTR}: This attribute holds the degree of membership of the value in \emph{labelTR} to the domain of wind
        energy.
    \item   \emph{synonymSet}: This attribute holds the list of English synonym phrases corresponding to this concept, if any.
    \item   \emph{membershipValueSynonymSet}: This attribute holds the list of the degrees of membership of the list of elements in the value
        of \emph{synonymSet} to the domain of wind energy.
    \item   \emph{synonymSetTR}: This attribute holds the list of English synonym phrases corresponding to this concept, if any.
    \item   \emph{membershipValueSynonymSetTR}: This attribute holds the list of the degrees of membership of the list of elements in the value
        of \emph{synonymSetTR} to the domain of wind energy.
\end{itemize}

As described above, the attributes beginning with \emph{membershipValue} hold weighted values corresponding to fuzzy membership functions, as
previously explained in \cite{kuccuk2015feeont}.

All of the 25 instances of OntoWind, shown in Figure \ref{fig:instances}, correspond to national or international organizations related to wind
energy. Apart from \emph{MGM} which models \emph{Turkish State Meteorological Service}, other instances include international organizations like
ECMWF (denoting \emph{European Centre for Medium-Range Weather Forecasts}) and WMO for \emph{World Meteorological Organization}, and national
ones like \emph{CENER} which is a Spanish institute for renewable energy research and \emph{NCAR} which corresponds to \emph{National Center for
Atmospheric Research} of the USA.

The final form of the improved and extended wind energy ontology, OntoWind, is made publicly-available at
\url{http://www.ceng.metu.edu.tr/~e120329/OntoWind.owl} as an OWL file. To summarize its characteristics again, OntoWind is an extended and
reorganized wind energy ontology which\footnote{We should note that the first two characteristics are not specific to OntoWind, as the first one
is a characteristic of both PQONT \cite{kucuk2010pqont} and FEEONT \cite{kuccuk2015feeont}, and the second one is a characteristic of FEEONT.}:

\begin{itemize}
    \item   supports multilinguality with its relevant attributes such as \emph{labelEN}, \emph{labelTR}, \emph{synonymSet}, and
        \emph{synonymSetTR}, and the ontology can be extended with similar attributes for other languages as previously pointed out,
    \item   supports attributes conveying crisp as well as weighted information (such as \emph{membershipValueLabel} and
        \emph{membershipValueLabelTR}) for the other attributes,
    \item   can also be considered as a larger knowledge base including the ontology concepts and their hierarchies as well as concept
        instances (as illustrated in Figure \ref{fig:instances}).
\end{itemize}

\section{Semantic System for the Surveillance of Wind Energy Information}\label{sec:system}

\begin{figure}
\center \scalebox{0.33}
{\includegraphics{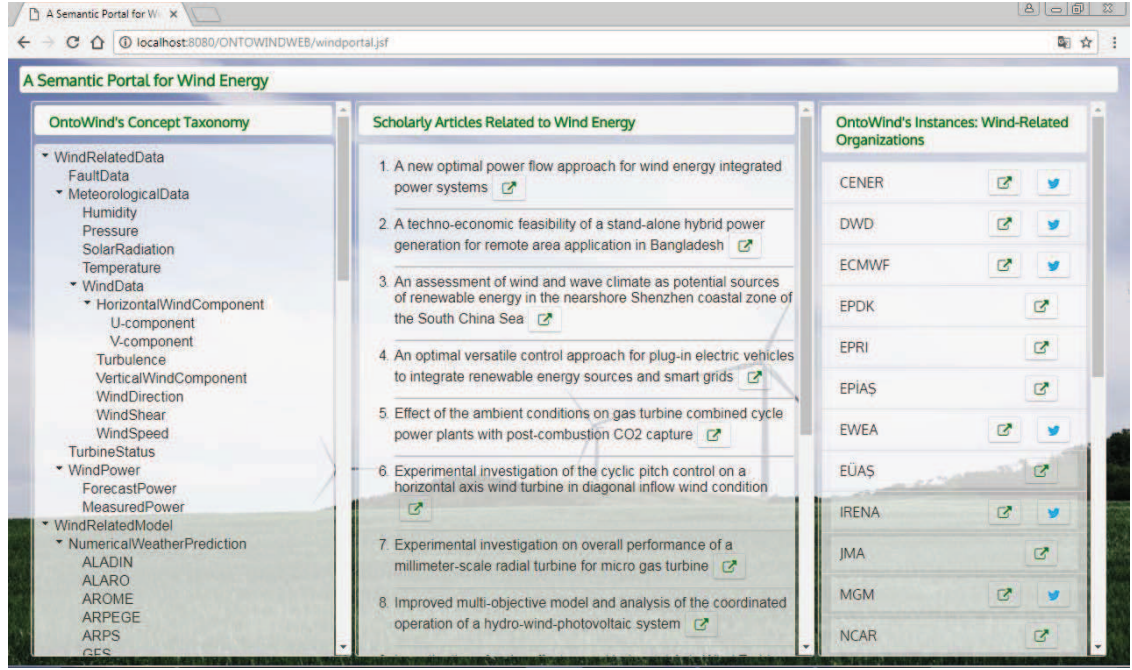}}\caption{A Snaphost of the Semantic Portal for Wind Energy Surveillance on the Web.}\label{fig:portal}
\end{figure}

In this section, we present a semantic portal for wind energy surveillance on the Web, based on OntoWind as the underlying source of semantic
information. The portal enables its users (i) to examine the concepts in OntoWind ontology, (ii) to view the wind-related research organizations
with links to their official Web sites and Twitter accounts, which correspond to the instances of OntoWind ontology, and (iii) to observe the
automatically-extracted wind-related scholarly articles. A snapshot of this Web-based semantic portal application is provided in Figure \ref{fig:portal}. The application is implemented using Java programming language, JavaServer Faces (JSF) technology \cite{jsf} and PrimeFaces JSF library \cite{primefaces} as the underlying technologies, and Java OWL API \cite{owl-api} to programmatically access OntoWind as an OWL file.

On the left panel of the portal, the taxonomy of the OntoWind concepts are provided in the form of a tree. The tree is automatically generated by
processing OntoWind using Java OWL API \cite{owl-api}. The users can freely examine all of the ontology concepts through this panel.

On the right panel of the portal, OntoWind's concept instances described in the previous section are listed. All of these instances are
significant organizations related to wind energy, that are automatically extracted from OntoWind. Based on the values of the \emph{webAddress}
and \emph{twitterAccount} attributes of these instances, related buttons to access the Web sites and official Twitter account pages of these
organizations automatically appear.

The final and the most significant feature of the semantic portal for wind energy is the panel appearing at the center of the portal page. This
panel presents automatically-extracted content from the Web, using OntoWind as the underlying semantic resource. In its current form, it
basically presents automatically-extracted scholarly articles related to wind energy. We have implemented this proof-of-concept extractor system,
we have followed the categorization strategy previously proposed in \cite{kuccuk2015feeont}. That is, we have built a text categorization system
based on OntoWind, which categorizes a given text as related to the wind energy domain, if it includes at least one concept from OntoWind and the
sum of the values of the \emph{membershipValueLabel} attributes corresponding to the included concepts is at least 1.0. Otherwise, the text is
categorized as irrelevant to the wind energy domain.

In order to test our categorizer, we have used 91 articles related to the energy domain which have been published in the 134th
volume\footnote{134th volume is the most recent volume of the journal, at the time of writing this paper.} of Elsevier's \emph{Energy} journal
\cite{elsevier}. For each journal article, we have executed our categorizer on a piece of text including the title, abstract, and keywords of
this article. We have built a similar categorizer for comparison purposes, which employs WONT instead of OntoWind, and this categorizer is also
executed on the same article texts. In order to create the answer key using which the performance rates of the categorizers will be calculated,
we have manually annotated each article in the set of 91 articles as relevant or irrelevant to the wind energy domain. At the end of this
process, it is found that 12 articles are related to wind energy while 79 articles are not. Based on this answer set, the performance evaluation
results of the OntoWind-based and WONT-based categorizers are presented in Table \ref{tab:sonuc}.

\begin{table}
  \caption{Evaluation Results of the OntoWind-Based and WONT-Based Classifiers.}
  \label{tab:sonuc}
    \centering
    \begin{tabular}{|l|c|c|c|c|c|}
    \hline
    &\emph{True}&\emph{False}&\emph{True}&\emph{False}&  \\
    &\emph{Positives}&\emph{Negatives}&\emph{Negatives}&\emph{Positives}&\emph{Accuracy}\\
    \hline
    OntoWind-Based &&&&& \\
    Categorizer & 10 & 2 & 75 & 4 & 93.4\% \\
    \hline
    WONT-Based &&&&& \\
    Categorizer & 10 & 2 & 64 & 15 & 81.3\% \\
    \hline
    \end{tabular}
\end{table}

When the results in Table \ref{tab:sonuc} are examined in details, we see that both the OntoWind-based and WONT-based categorizers are good at
finding the relevant articles. Yet, WONT-based categorizer spuriously outputs irrelevant articles as relevant as revealed with the high number of
false positives. In general, it can be concluded that the performance of the categorizer based on OntoWind is highly favorable and better when
compared with the results obtained by the WONT-based categorizer. This finding provides evidence for the coverage and utility of OntoWind in semantic applications.

We have previously stated that a similar categorization experiment has been reported in \cite{kuccuk2015feeont}. In the experiment given in
\cite{kuccuk2015feeont}, the data set contains articles from many different domains while in the current study, the articles are already from the
energy domain in general which makes it harder to extract those ones relevant to the specific domain of wind energy. This difference in the data
sets also explains the performance difference of the OntoWind-based categorizer of the current study and the categorizer presented in
\cite{kuccuk2015feeont} which achieves an accuracy of 99.09\%.

Our main objectives in this categorization experiment are (i) to showcase the practical contribution of the improved and extended OntoWind
ontology over its predecessor, WONT, which is revealed with the findings given in Table \ref{tab:sonuc} and (ii) to use the
automatically-extracted articles by the OntoWind-based categorizer in our semantic wind energy portal as shown in the middle panel of the portal
shown in Figure \ref{fig:portal}. The middle panel titled ``\emph{Scholarly Articles Related to Wind Energy}" also facilitates access to the
actual publisher pages of these articles through the accompanied links.

The semantic wind energy portal application is a proof-of-concept system and an initial prototype. The following extensions to this application
are envisaged as part of future work:

\begin{itemize}
  \item The portal will be extended to list recent automatically-extracted news articles related to wind energy, so that it will be a plausible
      information hub for semantic information related to wind energy.
  \item Similarly, recent social media content related to wind energy like relevant tweets will also be automatically extracted and presented
      through the portal.
  \item Currently, the semantic portal for wind energy is not publicly accessible. After the above mentioned extensions, we will make the
      portal accessible through the Internet.
\end{itemize}

As wind energy and renewable energy resources in general are important research topics with considerable public impact, we believe that both our
extended and improved ontology, OntoWind, and the semantic portal application for wind energy are important contributions to the related literature.

\section{Conclusion}\label{sec:conc}
Wind energy is a ubiquitous renewable energy type and several research topics regarding wind energy still need extensive efforts to fulfill the
research needs. One of these topics is the representation and application of semantic information regarding wind energy. In this study, we
present an extended and improved wind energy ontology, called OntoWind, in order to conveniently represent the semantic information in the wind
energy domain. We restructure, improve, and extend an existing semi-automatically created ontology for this domain by adding new concepts,
attributes, and instances to arrive at OntoWind. After making it publicly-available, we have developed a semantic portal for wind energy
utilizing the final form of the ontology, hence showed its contribution to a genuine semantic application. The portal currently presents
automatically extracted scholarly articles in addition to the concept taxonomy and instances of OntoWind. As part of future work, we plan to
extend the portal to publish other semantic content related to wind energy, like recent news articles and social media content. Other directions
of future work include similarly building ontologies for other renewable energy resources and then consolidating them to make them applicable to
significant semantic applications regarding renewable energy and energy in general.

\end{document}